\documentclass[letterpaper]{article} 
\usepackage{aaai23}  
\usepackage{times}  
\usepackage{helvet}  
\usepackage{courier}  
\usepackage[hyphens]{url}  
\usepackage{graphicx} 
\usepackage{natbib}  
\usepackage{caption} 
\usepackage{algorithm}
\usepackage{algorithmic}

\usepackage{amsmath}
\usepackage{bm}
\usepackage{amssymb}
\usepackage{multirow}
\usepackage{graphicx}
\usepackage{subfigure}
\usepackage{capt-of}
\usepackage{booktabs}

\usepackage{newfloat}
\usepackage{listings}
\DeclareCaptionStyle{ruled}{labelfont=normalfont,labelsep=colon,strut=off} 
\floatstyle{ruled}
\newfloat{listing}{tb}{lst}{}
\floatname{listing}{Listing}
\title{Boosting Few-Shot Text Classification via Distribution Estimation}
\author {
    Han Liu\textsuperscript{\rm 1},
    Feng Zhang\textsuperscript{\rm 2},
    Xiaotong Zhang\textsuperscript{\rm 1}\thanks{Corresponding author.},
    Siyang Zhao\textsuperscript{\rm 1},
    Fenglong Ma\textsuperscript{\rm 3},\\
    Xiao-Ming Wu\textsuperscript{\rm 4},
    Hongyang Chen\textsuperscript{\rm 5},
    Hong Yu\textsuperscript{\rm 1},
    Xianchao Zhang\textsuperscript{\rm 1}
}
\affiliations {
    \textsuperscript{\rm 1} Dalian University of Technology, Dalian, China\\
    \textsuperscript{\rm 2} Peking University, Beijing, China\\
    \textsuperscript{\rm 3} The Pennsylvania State University, Pennsylvania, USA\\
    \textsuperscript{\rm 4} The Hong Kong Polytechnic University, Hong Kong, China\\
    \textsuperscript{\rm 5} Zhejiang Lab, Hangzhou, China\\
    liu.han.dut@gmail.com, zfeng.maria@gmail.com, zxt.dut@hotmail.com, zhao\_siyang@mail.dlut.edu.cn, \\
    fenglong@psu.edu, csxmwu@comp.polyu.edu.hk, dr.h.chen@ieee.org, \{hongyu,xczhang\}@dlut.edu.cn
}

\usepackage{bibentry}

\begin{document}

\maketitle

\begin{abstract}
Distribution estimation has been demonstrated as one of the most effective approaches in dealing with few-shot image classification, as the low-level patterns and underlying representations can be easily transferred across different tasks in computer vision domain. However, directly applying this approach to few-shot text classification is challenging, since leveraging the statistics of known classes with sufficient samples to calibrate the distributions of novel classes may cause negative effects due to serious category difference in text domain. To alleviate this issue, we propose two simple yet effective strategies to estimate the distributions of the novel classes by utilizing unlabeled query samples, thus avoiding the potential negative transfer issue. Specifically, we first assume a class or sample follows the Gaussian distribution, and use the original support set and the nearest few query samples to estimate the corresponding mean and covariance. Then, we augment the labeled samples by sampling from the estimated distribution, which can provide sufficient supervision for training the classification model. Extensive experiments on eight few-shot text classification datasets show that the proposed method outperforms state-of-the-art baselines significantly.
\end{abstract}

\section{Introduction}
Text classification plays a fundamental and crucial role in natural language processing, which has been widely applied to various real applications, such as intent detection \cite{DBLP:conf/coling/LouvanM20}, sentiment analysis \cite{DBLP:journals/nca/KumarA21}, news classification \cite{DBLP:conf/icwsm/BozarthB20} and so on.
Traditional text classification methods \cite{DBLP:conf/acl/JohnsonZ17, DBLP:conf/naacl/DevlinCLT19} have achieved impressive performance, which require a large amount of
labeled instances per class for training. However, collecting and annotating sufficient data is a time-consuming and labor-intensive process, sometimes even unachievable in industry, which motivates few-shot text classification. 

Few-shot learning is a paradigm for solving the data
scarcity issue, which aims to detect novel categories with very limited labeled examples by using prior knowledge learned from known categories. Several kinds of methods have been proposed to meet this challenge. Meta-learning based methods aim to train a generalized model which can quickly adapt to new tasks \cite{finn2017model, mann, snell2017prototypical,DBLP:conf/emnlp/LiuZ0ZZ21,DBLP:conf/kdd/LiuZ0ZS0Z22}. This type of methods has been successfully applied to solve the few-shot learning problem. Fine-tuning based methods usually train a model on the base set first and then transfer to novel classes via adjusting the model parameters \cite{howard-ruder-2018-universal, gururangan-etal-2020-dont}, which are susceptible to the overfitting problem. Their variants like prompt-based and entailment-based methods \cite{lmbff,efl} can mitigate the above issue and have achieved promising performance. It is worth noting that most previous works focus on developing stronger models, but less attention has been paid to the property of the data itself. Intuitively, when more informative data is available for supervision, the model tends to generalize well during evaluation.

In order to explore the problem from the perspective of data itself, several data-augmentation based methods \cite{DBLP:conf/acl-deeplo/KumarGLC19,ProtAugment,ContrastNet} have been proposed. However, these methods require the design of a complex model and loss function to learn how to generate examples. Recently, one variant of data-augmentation based methods named distribution calibration has shown to be effective in few-shot image classification. It first estimates the distribution of the unseen classes by transferring statistics from the seen classes, and then samples an adequate number of examples to expand the size of labeled data. Nevertheless, this method cannot directly extend to text domain. The main reason is as follows. In vision domain, low-level patterns and their corresponding representations can usually be shared across classes. For example the classes \emph{white\_wolf} and \emph{arctic\_fox} from ImageNet \cite{DBLP:conf/cvpr/DengDSLL009} are very similar. The category difference, however, tends to be serious in text domain. For example, the classes \emph{get\_weather} and \emph{play\_music} from SNIPS \cite{snips} are entirely different. That is to say, the unseen classes probably have no overlap with the seen classes in text domain. Simply transferring distribution statistics from seen data seems not a good solution, as some distribution statistics from seen classes may be biased or even harmful to the unseen classes.

\begin{figure*}
	\centering
	\includegraphics[height=4.6cm]{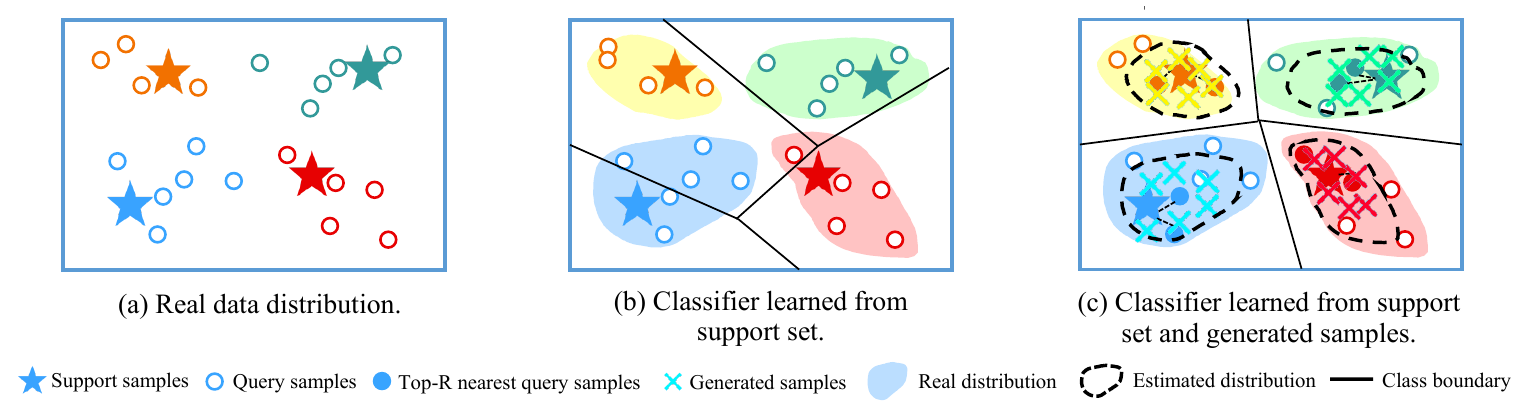}
	\caption{Illustration of a simple 4-way 1-shot task. Figure (a) shows the real data distribution, which contains one labeled support sample per class and several unlabeled query samples. Figure (b) shows that the classifier learned from only one support sample may cause the serious overfitting issue, and the class boundary is biased. Figure (c) shows the classifier learned from support set and generated samples based on our estimated distribution, which has better class-discriminative ability.}
	\label{framework}
\end{figure*}

In this paper, we propose two simple yet effective strategies (way-based and shot-based strategies) to estimate the distributions of the novel classes by exploiting unlabeled query samples instead of adequate samples from seen classes, thus circumventing the possible adverse impact caused by serious category difference. In particular, we assume a class or sample obeys the Gaussian distribution, and use the original support set and the nearest several query samples to estimate the corresponding mean and covariance. Based on the approximated distribution, we generate a sufficient amount of labeled data to augment the support set, thus boosting the model performance. Figure 1 gives a 4-way 1-shot task to illustrate the drawbacks of previous methods and the advantages of our proposed strategies. To verify the effectiveness of the proposed methods, we conduct extensive experiments on eight public datasets. The empirical results show that the proposed strategies can achieve promising results compared with other strong baselines.
 
\section{Related Work}
\subsection{Meta-Learning Based Methods}
Meta-learning aims to design a model which can well adapt or generalize to new tasks and new environments that have never been encountered with only a few training examples. Existing meta-learning based methods can be divided into three types. (1) Optimization-based methods, such as MAML \cite{finn2017model} and Reptile \cite{DBLP:journals/corr/abs-1803-02999}, intend to learn how to optimize the gradient descent procedure so that the model can be effective in learning with a few instances. (2) Model-based methods, such as MANN \cite{mann} and MetaNet \cite{metanet}, rely on the modules that can update the parameters rapidly and effectively with a few steps. (3) Metric-based methods like matching network \cite{matchingnet}, prototypical network \cite{snell2017prototypical}, relation network \cite{relationnet} and induction network \cite{DBLP:conf/emnlp/GengLLZJS19}, first learn an embedding space, and then use a metric to classify new category cases based on the proximity to labeled examples.

\subsection{Fine-Tuning Based Methods}
Traditional fine-tuning algorithms usually use a few samples belonging to the unseen classes to update the parameters of the models pre-trained on the seen classes with adequate samples, which is a straightforward way to deal with few-shot learning. However, these algorithms inevitably suffer from the over-fitting issue due to data scarcity. To mitigate
this issue, \citet{howard-ruder-2018-universal} and \citet{gururangan-etal-2020-dont} propose to train the models with the language model objective function on task-specific unlabeled data before fine-tuning models on the target task. \citet{phang2018sentence} propose to train the model with data-rich intermediate supervised tasks before fine-tuning it on the target task. Recently, prompt-based and entailment-based methods seem potential in dealing with the few-shot learning task. LM-BFF \cite{lmbff} introduces automatic prompt generation and incorporates the demonstrations as additional context to fine-tune smaller language models on a handful of annotated examples. EFL \cite{efl} reformulates NLP tasks as textual entailment instead of cloze questions, and provides fine-grained label-specific descriptions instead of a single task description, thus achieving promising performance.

\subsection{Data Augmentation Based Methods}
Data augmentation is a tried and true method to solve the data sparsity problem. Conventional augmentation methods focus on word substitution \cite{NIPS2015_250cf8b5}. EDA \cite{eda} proposes four simple operations, synonym replacement, random insertion, random swap, and random deletion. Recently, some strong methods are specifically proposed for few-shot text classification. \citet{DBLP:conf/acl-deeplo/KumarGLC19} explore six feature space data augmentation methods to improve performance in few-shot intent classification. PROTAUGMENT \cite{ProtAugment} introduces a short-text paraphrasing model that produces diverse paraphrases of the original text as data augmentation. ContrastNet \cite{ContrastNet} leverages data augmentation to train the supervised contrastive representation model under the regularization of a task-level unsupervised contrastive loss and an instance-level unsupervised contrastive loss. Recently, distribution estimation \cite{freelunch} has shown to be powerful in dealing with few-shot image classification, which first calibrates the distribution of the unseen classes by transferring statistics from the seen classes. Then an adequate number of examples are sampled from the calibrated distribution to expand the inputs to the classifier. Obviously, its core goal is to generate more samples based on the estimated distribution, thus providing more supervision for training the classification model. However, it heavily relies on the strong assumption that there always exist seen classes that are similar to an unseen class, which probably not holds in text domain.

\section{The Proposed Method}
\subsection{Problem Formulation}
In this paper, we use the episode learning strategy to explore few-shot text classification. Specifically, the data is divided into two parts: seen class set $\mathcal{C}_\text{seen}$ and unseen class set $\mathcal{C}_\text{unseen}$, and $\mathcal{C}_\text{seen} \bigcap \mathcal{C}_\text{unseen}={\emptyset}$. A classifier is trained with numerous samples from $\mathcal{C}_\text{seen}$, and it is quickly adopted to $\mathcal{C}_\text{unseen}$ with only a few labeled data from $\mathcal{C}_\text{unseen}$. Meta learning provides an effective solution for few-shot learning, which commonly follows the $N$-way $K$-shot setting, i.e., for each task, there are $N$ classes and each class has $K$ supports (labeled samples). In general, meta-learning contains two phases: training and testing.

In the training phase, the meta-classifier is trained on $N_{train}$ tasks. In each training task, it consists of a support set and a query set. To construct the train task, $N$ classes are randomly picked from $\mathcal{C}_\text{seen}$. The support set is composed of randomly selecting $K$ labeled samples from each of the $N$ classes, i.e., $\mathcal{S}=\{(\bm x_i, y_i)\}_{i=1}^m$, where $\bm x_i$ is a data sample, $y_i$ is the class label and $m=N\times K$. The query set consists of a portion of the remaining samples from these $N$ classes, i.e., $\mathcal{Q}=\{(\bm x_j, y_j)\}_{j=1}^n$, where $n$ is the number of queries.

In the testing phase, the trained meta-classifier is used to predict the labels of queries in $N_{test}$ tasks. In each testing task, it also has a support set and a query set. In a similar manner, $N$ classes are randomly sampled from the test classes $\mathcal{C}_\text{unseen}$. The support set and query set are constructed in the same way as those in the meta-training phase. As the labels of queries are unknown in testing stage, the query set in test task can be represented as $\mathcal{Q}=\{\bm x_j\}_{j=1}^n$. The goal is to predict the class labels for these queries.

\subsection{Basic Few-Shot Classifier}\label{basic}
We take a popular metric-based meta-learning method -- prototypical network \cite{snell2017prototypical} -- as the basic few-shot classifier. The core idea of prototypical network is to learn a mapping (metric) $\phi$ that projects support and query samples into an embedding space, and then classify the queries by learning their relations according to the Euclidean distance in that space. Specifically, for each training task, the prototype $\bm P^c$ of the $c$-th ($c=1, 2, \ldots, N$) class is obtained by averaging $K$ mapped supports ${\phi(\bm x_i^c)}$ in this class, i.e., $\bm P^c = \frac1K\sum_{i=1}^K \phi(\bm x_i^c)$. For a query $\bm x_q$, the probability of $\bm x_q$ belonging to the $c$-th class is computed by a softmax function with the Euclidean distances between ${\phi(\bm x_q)}$ and the prototypes:
\begin{equation}\label{PN1}
  f_{qc} = \frac{\exp(-||\phi(\bm x_q) - \bm P^c||_2^2)}{\sum_{i=1}^N \exp(-||\phi(\bm x_q) - \bm P^i||_2^2)},
\end{equation}
where the mapping $\phi$ is learned by minimizing the cross-entropy loss. Formally,
\begin{equation} \label{PN2}
\mathcal{L}_{basic} = \min_{\phi} \sum_{q=1}^{n}\sum_{c=1}^N- y_{qc} \log f_{qc},
\end{equation}
where $y_{qc}=1$ if $\bm x_q$ belongs to the $c$-th class, otherwise $y_{qc}=0$. $n$ is the number of queries.

\subsection{Distribution Estimation}
Distribution calibration \cite{freelunch} attempts to calibrate the distributions of unseen classes with few samples by transferring statistics from seen classes with sufficient samples in vision domain. This method heavily relies on the strong assumption that there always exist seen classes which are similar with an unseen class. However, this assumption does not always hold well in text domain. To tackle this issue, we propose two simple yet effective distribution estimation strategies by utilizing unlabeled query samples.

Considering an $N$-way $K$-shot task, given a novel class $c$, its $K$ support samples can be represented as $\{\bm{x}_1, ..., \bm{x}_K\}$. For each $\bm{x}_i$, we can calculate the top $R$ nearest query samples of $\bm{x}_i$ according to the Euclidean distance in the embedding (mapping) space, and we denote this set as $\{\bm{a}_{i1}, ..., \bm{a}_{iR}\}$. Here $R$ is a hyperparameter.

\subsubsection{Way-Based Distribution Estimation} For the way-based distribution estimation strategy, we treat each way (class) as a random variable which follows the Gaussian distribution in the embedding space. In general, the mean of the Gaussian distribution can be obtained by averaging the embedding of each sample in support set:
\begin{equation}
	\boldsymbol{\mu}_s = \frac{1}{K}\sum_{i=1}^{K} \phi(\bm x_i),
\end{equation}
where $\phi$ is the feature extraction function.

In order to better estimate the distribution of the novel class, we attempt to use the top $R$ query samples to calibrate the estimation result. Specifically, we first calculate the mean of $\{\bm{a}_{11},..., \bm{a}_{1R},..., \bm{a}_{K1}, ..., \bm{a}_{KR}\}$ with:
\begin{equation}
	\boldsymbol{\mu}_q = \frac{1}{{KR}}\sum_{i=1}^{K} \sum_{j=1}^{R}\phi(\bm{a}_{ij}).
\end{equation}
Then the final estimated mean $\bm{\mu}_{way}$ can be represented as:
\begin{equation}\label{way}
\begin{split}
	\bm{\mu}_{way} & =  \frac{1}{2}(\boldsymbol{\mu}_s+\boldsymbol{\mu}_q)\\
	 & = \frac{1}{2K}\sum_{i=1}^{K} \phi(\bm{x}_i) + \frac{1}{2KR} \sum_{i=1}^{K} \sum_{j=1}^{R} \phi(\bm{a}_{ij}).
\end{split}
\end{equation}

In a similar manner, we can estimate the covariance matrix $\bm{\Sigma}_{way}$ of the Gaussian distribution with:
\begin{equation}
	\bm{\Sigma}_{way} = \frac{1}{2}(\bm{\Sigma}_{s}+\bm{\Sigma}_{q}),
\end{equation}
where $\bm{\Sigma}_{s} \in \mathcal{R}^{d \times d}$  and $\bm{\Sigma}_{q} \in \mathcal{R}^{d \times d}$ can be calculated with:
\begin{equation}
	\bm{\Sigma}_{s} = \frac{1}{K - 1} \sum_{i=1}^{K}(\phi(\bm x_i) - \boldsymbol{\mu}_s )(\phi(\bm x_i) - \boldsymbol{ \mu}_s )^T,
\end{equation}
\begin{equation}
	\bm{\Sigma}_{q} = \frac{1}{KR - 1} \sum_{i=1}^{K}\sum_{j=1}^{R}(\phi(\bm{a}_{ij}) - \boldsymbol{\mu}_q )(\phi(\bm{a}_{ij}) - \boldsymbol{ \mu}_q )^T.
\end{equation}

\subsubsection{Shot-Based Distribution Estimation} For the shot-based distribution estimation strategy, we follow \cite{freelunch} to treat each shot (support sample) as a random variable which obeys the Gaussian distribution. For each support sample $\bm x_i$, as it can represent the original mean, we only need to use the top $R$ query samples to adjust it. Specifically, the estimated mean of the support sample $\bm x_i$ can be obtained by:
\begin{equation}
	\bm{\mu}_i = \frac{1}{2}(\phi (\bm x_i)+\frac{1}{R}\sum_{j=1}^{R}\phi(\bm{a}_{ij})),
\end{equation}
and the estimated covariance matrix $\bm{\Sigma}_{i}$ of the support sample $\bm x_i$ can be calculated by:
\begin{equation}
	\bm{\Sigma}_i = \frac{1}{R-1} \sum_{j=1}^{R} (\phi(\bm{a}_{ij}) - \bm{\mu}_i)(\phi(\bm{a}_{ij}) - \bm{\mu}_i)^T.
\end{equation}
By using the above distribution estimation method, for a class $c$ with $K$ support samples, its distribution can be represented as the set $\{\mathcal{N}(\boldsymbol{\mu}_1, \bm{\Sigma}_1), ...,  \mathcal{N}(\boldsymbol{\mu}_K, \bm{\Sigma}_K)\}$.

\subsection{Distribution Sampling}
According to the estimated distribution, we can generate more samples which can provide sufficient supervision for training the classification model.

\subsubsection{Way-Based Distribution Sampling} Given an unseen class $c$, in this scenario we can generate the samples with label $c$ by sampling from the following Gaussian distribution:
\begin{equation}
    \mathbb{D}_c = \{(\bm{x}, c) | \bm{x} \sim \mathcal{N}(\boldsymbol{\mu}_{way}, \bm{\Sigma}_{way})\}.
\end{equation}
After generating a series of samples, we can combine the original support set and the generated samples together to serve as the training data for a task.

\subsubsection{Shot-Based Distribution Sampling} Given an unseen class $c$, we denote $\mathbb{S}_c = \{(\boldsymbol{\mu}_1, \bm{\Sigma}_1), ...,  (\boldsymbol{\mu}_K, \bm{\Sigma}_K)\}$ as the set of mean and covariance pairs. We can generate the samples with label $c$ by sampling from the following Gaussian distribution:
\begin{equation}
	\mathbb{D}_c = \{(\bm{x}, c) | \bm{x} \sim \mathcal{N}(\boldsymbol{ \mu}, \bm{\Sigma}), \forall (\boldsymbol{ \mu}, \bm{\Sigma}) \in \mathbb{S}_c\}.
\end{equation}
After the sampling procedure, we can train the whole model with the original support set and the generated samples.

\subsubsection{Relationship between Way-Based and Shot-Based Strategies}
Considering the shot-based distribution sampling, if we sample uniformly from the distribution $\mathbb{D}_c = \{(\bm{x}, c) | \bm{x} \sim \mathcal{N}(\boldsymbol{ \mu}, \bm{\Sigma}), \forall (\boldsymbol{ \mu}, \bm{\Sigma}) \in \mathbb{S}_c\}$, we can get the overall mean $\bm{\mu}_{shot}$ can be represented as:
\begin{equation}\label{relation}
\begin{split}
	\bm{\mu}_{shot} & = \frac{1}{K}\sum_{i=1}^{K}\bm{\mu}_i \\
& = \frac{1}{K}\sum_{i=1}^{K}(\frac{1}{2}(\phi(\bm x_i)+\frac{1}{R}\sum_{j=1}^{R}\phi(\bm{a}_{ij})))\\
& = \frac{1}{2K}\sum_{i=1}^{K} \phi(\bm{x}_i) + \frac{1}{2KR} \sum_{i=1}^{K} \sum_{j=1}^{R} \phi(\bm{a}_{ij}).
\end{split}
\end{equation}
From Eq. (\ref{way}) and Eq. (\ref{relation}), it is easy to observe that the way-based and shot-based distribution estimation strategies share the same mean, which indicates that these two estimated distributions probably overlap heavily. In addition, in 
the extreme 1-shot scenario, way-based and shot-based distribution estimation methods are equivalent.

\subsection{Training and Testing Phases}
\subsubsection{Training}
During the training phase, we use the prototypical network loss and the generation loss simultaneously. For the prototypical network loss $\mathcal{L}_{basic}$, when calculating the prototype for each class, we combine the original support set and the generated samples as the final support set, and the remaining calculation process can refer to Eqs. (\ref{PN1}) and (\ref{PN2}). In terms of the generation loss, it aims to guarantee that the generated samples are close to their original center and away from other centers, thus improving the confidence of generated samples. To achieve this goal, the generation loss can be written as:
\begin{equation}
	\mathcal{L}_{gen} = \frac{1}{|\mathbb{D}|} \sum_{(x_*, y_*) \in \mathbb{D}} - \log p( y = y_* | x_*, \mathcal{S}),
\end{equation}
where $\mathbb{D} =  \cup_{c} \mathbb{D}_c$ is the overall generated data, and $\mathcal{S}$ is the original support set. Then the overall loss function is:
\begin{equation}
	\mathcal{L}_{total} = \mathcal{L}_{basic}+ \lambda \mathcal{L}_{gen},
\end{equation}
where $\lambda$ is a trade-off hyperparameter. By minimizing $\mathcal{L}_{total}$ with gradient descent methods, all trainable model parameters can be learned.
\subsubsection{Testing}
In the testing phase, given an $N$-way $K$-shot task, we first estimate the distribution with way-based or shot-based approaches. Based on the estimated distribution, we generate the corresponding samples and combine them with the original support set as the final support set. Finally, we predict the class label for each query sample by the prototypical network.

\section{Experiments}

\begin{table*}[t]
	\centering
	\begin{tabular}{lccc}
		\toprule
		\textbf{Dataset} & \textbf{\#samples} & \textbf{Avg. text length} & \textbf{\#train / valid / test (total) classes} \\ \midrule
		HuffPost \cite{iclr/BaoWCB20}   & 36900             & 11.48           & 20 / 5 / 16 (41)      \\
		Amazon \cite{amazon}            & 24000             & 143.46          & 10 / 5 / 9 (24)       \\
		Reuters \cite{iclr/BaoWCB20}    & 620               & 181.41          & 15 / 5 / 11 (31)      \\
		20News \cite{20news}            & 18828             & 279.32          & 8 / 5 / 7 (20)        \\
\midrule
	    Banking77 \cite{banking}   & 13083             & 11.77        & 25 / 25 / 27 (77)          \\
		HWU64 \cite{hwu}           & 11036             & 6.57         & 23 / 16 / 25 (64)          \\
		Liu57 \cite{hwu}           & 25478             & 6.66         & 18 / 18 / 18 (54)          \\
		Clinic150 \cite{oos}       & 22500             & 8.31         & 50 / 50 / 50(150)          \\
	 \bottomrule
	\end{tabular}
	\caption{Dataset statistics.}
	\label{statistics}
\end{table*}

\subsection{Datasets}
We follow \cite{ContrastNet} to conduct experiments on eight text classification datasets, including four intent detection datasets: Banking77, HWU64, Clinic150, and Liu57, and four news or review classification datasets: HuffPost, Amazon, Reuters, and 20News. The average length of sentences in news or review classification datasets is much longer than those in intent detection datasets. Table  \ref{statistics} concludes the statistics of all datasets.

\textbf{(1) HuffPost} \cite{iclr/BaoWCB20} is a news headlines dataset with 41 classes, which are published on HuffPost from 2012 to 2018. They are shorter and less grammatical than formal sentences.

\textbf{(2) Amazon} \cite{amazon} consists of 142.8 million customer reviews from 24 product categories. Following \cite{mlada}, we use a subset, having 1000 reviews per category.

\textbf{(3) Reuters} \cite{iclr/BaoWCB20} is collected shorter Reuters articles in 1987. Following \cite{iclr/BaoWCB20}, we discard multi-label articles and only use 31 classes, having at least 20  articles.

\textbf{(4) 20News} \cite{20news} covers 18828 documents from news discussion forums under 20 topics.

\textbf{(5)  Banking77} \cite{banking} is a fine-grained single-domain dataset for intent detection. It consists of 13083 customer service queries labeled with 77 intents, in which some categories are similar and may have overlap with others.

\textbf{(6) HWU64} \cite{hwu} contains 11036 utterances covering 64 intents in 21 domains. The examples are from a real-world home robot, with multi-domain utterances, e.g., email, music, weather and so on.

\textbf{(7) Liu57} \cite{hwu} is composed of 25478 user utterances from 54 classes. The dataset is collected from Amazon Mechanical Turk.

\textbf{(8) Clinic150} \cite{oos} contains 150 intents and 23700 examples across 10 domains. It has 22500 user utterances evenly distributed in every intent and 1200 out-of-scope queries. Here we ignore these out-of-scope examples.

\begin{table*}[t]
	\centering
	\setlength{\tabcolsep}{2.2mm}{
		\begin{tabular}{lcccccccccc}
			\toprule
			\multirow{2}{*}{\textbf{Method}} & \multicolumn{2}{c}{\textbf{HuffPost}} & \multicolumn{2}{c}{\textbf{Amazon}} & \multicolumn{2}{c}{\textbf{Reuters}} & \multicolumn{2}{c}{\textbf{20News}} & \multicolumn{2}{c}{\textbf{Average}} \\ \cmidrule{2-11}
			& 1-shot & 5-shot  & 1-shot & 5-shot  & 1-shot & 5-shot  & 1-shot & 5-shot  & 1-shot & 5-shot  \\ \midrule
			Prototypical Networks  & 35.7  & 41.3 & 37.6 & 52.1 & 59.6 & 66.9 & 37.8  & 45.3  & 42.7  & 51.4  \\
			MAML                   & 35.9  & 49.3 & 39.6 & 47.1 & 54.6 & 62.9 & 33.8  & 43.7  & 40.9  & 50.8  \\
			Induction Networks     & 38.7  & 49.1 & 34.9 & 41.3 & 59.4 & 67.9 & 28.7  & 33.3  & 40.4  & 47.9  \\
			HATT                   & 41.1  & 56.3 & 49.1 & 66.0 & 43.2 & 56.2 & 44.2  & 55.0  & 44.4  & 58.4  \\
			DS-PSL                 & 43.0  & 63.5 & 62.6 & 81.1 & 81.8 & \textbf{96.0}& 52.1  & 68.3  & 59.9 & 77.2  \\
			MLADA                  & 45.0  & 64.9 & 68.4 & 86.0 & 82.3 & \textbf{96.7}& 59.6  & 77.8  & 63.9 & 81.4  \\
			ContrastNet            & 51.8  & 67.8 & 73.5 & 83.6 & 88.5 & 94.6 & 70.9  & 80.5  & 71.2  & 81.6   \\
			TPN	                   & 50.6  & 69.5 & 76.0 & 84.9	&\textbf{91.4}	&93.1	&63.0	&69.4	&70.3	&79.2 \\\midrule
			DC                     & 47.7  & 67.0 & 70.6 & 84.2 & 84.9 & 93.8 & 65.6  & 79.6  & 67.2  & 81.2    \\
			DC-DE                  & 49.2  & 68.3 & 73.9 & 85.0 & 88.7 & 94.2 & 68.8  & 80.9  & 70.2  & 82.1    \\\midrule
            Shot-DE (Ours)          & \textbf{51.9} & \textbf{71.4} & \textbf{76.1} & \textbf{86.9} & \textbf{90.6} & 95.1 & \textbf{71.0} & \textbf{83.2} & \textbf{72.4}  & \textbf{84.2}  \\
			Way-DE (Ours)         & \textbf{51.9} & \textbf{71.7} & \textbf{76.1} & \textbf{87.4} & \textbf{90.6} & 95.2 & \textbf{71.0} & \textbf{83.2} & \textbf{72.4}  & \textbf{84.4}  \\	\bottomrule
	\end{tabular}}
	\caption{The 5-way 1-shot and 5-shot average accuracy on  news or review classification datasets.}
	\label{text}
\end{table*}

\begin{table*}[h]
	\centering
		\setlength{\tabcolsep}{2mm}{
			\begin{tabular}{lcccccccccc}
				\toprule
				\multirow{2}{*}{\textbf{Method}} & \multicolumn{2}{c}{\textbf{Banking77}} & \multicolumn{2}{c}{\textbf{HWU64}} & \multicolumn{2}{c}{\textbf{Liu57}} & \multicolumn{2}{c}{\textbf{Clinic150}} & \multicolumn{2}{c}{\textbf{Average}} \\ \cmidrule{2-11}
				& 1-shot  & 5-shot  & 1-shot  & 5-shot  & 1-shot  & 5-shot  & 1-shot   & 5-shot   & 1-shot   & 5-shot   \\ \midrule
				PROTAUGMENT           & 86.9  & 94.5  & 82.4  & 91.7  & 84.4   & 92.6   & 94.9  & 98.4   & 87.2   & 94.3 \\
				PROTAUGMENT (bigram)  & 88.1  & 94.7  & 84.1  & 92.1  & 85.3   & 93.2   & 95.8  & 98.5   & 88.3   & 94.6 \\
				PROTAUGMENT (unigram) & 89.6  & 94.7  & 84.3  & 92.6  & 86.1   & 93.7   & 96.5  & 98.7   & 89.1   & 94.9 \\
				ContrastNet           & \textbf{91.2} & \textbf{96.4} & 86.6   & 92.6   & 85.9  & 93.7   & 96.6   & 98.5  & 90.1  & 95.3 \\
				TPN	                  & 90.4  & 94.8  & 83.7  & 91.5  & 86.6   & 93.2   & 97.1  & 98.1   & 89.5   & 94.4 \\\midrule
				DC                    & 86.8  & 94.9  & 79.4  & 90.7  & 84.8   & 92.9   & 95.5  & 98.6   & 86.6   & 94.3 \\
				DC-DE                 & 88.9  & 95.1  & 85.3  & 92.8  & 88.2   & 94.0   & \textbf{98.8}  & 99.0   & 90.3  & 95.2 \\\midrule	
			    Shot-DE (Ours)         & \textbf{90.5}  & \textbf{95.8}  & \textbf{87.1}  & \textbf{93.5}  & \textbf{90.4}  & \textbf{95.2} & \textbf{98.0} & \textbf{99.2}  & \textbf{91.5}  & \textbf{95.9} \\
			    Way-DE (Ours)  & \textbf{90.5}  & 95.4  & \textbf{87.1}  & \textbf{93.4}  & \textbf{90.4}  & \textbf{95.5} & \textbf{98.0}  & \textbf{99.3}  & \textbf{91.5}  & \textbf{95.9} \\\bottomrule
		\end{tabular}}
		\caption{The 5-way 1-shot and 5-shot average accuracy on intent detection datasets.}
		\label{intent}
\end{table*}

\subsection{Baselines}
We compare the proposed way-based distribution estimation (Way-DE) and shot-based distribution estimation (Shot-DE) with the following strong baselines. 

\textbf{(1) Prototypical Network} \cite{snell2017prototypical} is a metric-based method which  calculates the prototype for
each class by averaging the corresponding support samples, and utilizes the negative Euclidean distance between query samples and prototypes to do the few-shot classification task.

\textbf{(2) MAML} \cite{finn2017model} is an optimization-based method, which learns a good model initialization, and adapts to new tasks by a small number of gradient steps.

\textbf{(3) Induction Network} \cite{DBLP:conf/emnlp/GengLLZJS19} leverages the dynamic routing algorithm to learn generalized class-wise representations.

\textbf{(4) HATT} \cite{DBLP:conf/aaai/GaoH0S19} is a hybrid attention-based prototypical network, which improves the model robustness greatly.

\textbf{(5) DS-FSL} \cite{iclr/BaoWCB20} is a framework to map distributional signatures into attention scores, thus guiding the fast adaptation to new categories.

\textbf{(6) MLADA} \cite{mlada} is a meta-learning adversarial domain adaptation network, which aims to improve the adaptive ability and generate generalized text embeddings for new classes.

\textbf{(7) ContrastNet} \cite{ContrastNet}  trains the supervised contrastive representation model
under the regularization of a task-level unsupervised contrastive loss and an instance-level unsupervised contrastive loss, which can prevent overfitting and generate better representations.

\textbf{(8) TPN} \cite{tpn} intends to learn to propagate labels from labeled support samples to unlabeled query samples via episodic training and a specific graph construction, which is a powerful transductive few-shot learning method.

\textbf{(9) DC} \cite{freelunch} calibrates novel class distribution using statistics from the seen classes with abundant samples based on similarity. 

\textbf{(10) DC-DE} is a variant of DC, which considers the statistics from seen classes and query data simultaneously to estimate the distribution. It is a baseline for validating whether seen classes may bring some side effects on performance.

\textbf{(11) PROTAUGMENT} \cite{ProtAugment} is an extension of Prototypical Network \cite{snell2017prototypical} using diverse paraphrasing data augmentation. It conducts an instance-level unsupervised loss on the vanilla prototypical network. PROTAUGMENT (unigram) and PROTAUGMENT (bigram) are two enhanced versions using different words paraphrasing strategies. 

Note that PROTAUGMENT is a method specifically-designed for intent detection, which is not suitable for long text classification, so we do not compare it in the news or review classification task. In addition, in the intent detection task, due to space and time limitation, we just compare with the most effective methods.

\subsection{Implementation Details}

\subsubsection{Evaluation Metric}  
We follow \cite{ContrastNet} to use the accuracy to evaluate the performance. All reported results are from 5 different runs, and in each run the training, validation and testing classes are randomly resampled.

\subsubsection{Parameter Settings}
We follow \cite{ContrastNet} to conduct experiments on 5-way 1-shot and 5-shot setting. In news and review classification task, we report the average accuracy on 1000 episodes sampled from test set, where the number of query instances per class in each episode is 25. In intent detection task, we report the average accuracy on 600 episodes sampled from test set for 4 intent detection datasets,  where the number of query instances per class in each episode is 5. In terms of feature extraction, for the news or review classification task, we use the pure pre-trained bert-base-uncased model. For the intent detection task, we use the further pre-trained BERT language model provided in \cite{ProtAugment}. We set $R = 10$ for the news or review classification task, while $R = 4$ for the intent detection task. For the loss function, we set $\lambda=0.1$, and optimize the model parameters using AdamW \cite{iclr/LoshchilovH19} with the initial learning rate $0.00001$ and dropout rate $0.1$. During distribution sampling, in 1-shot and 5-shot scenarios, we generate 20 and 100 samples per class respectively. All the hyper-parameters are selected based on the performance of the validation set.

\begin{figure*}[t]
	\centering
	\subfigure[Support and query examples.]{
		\includegraphics[height=3.6cm,width=5.6cm]{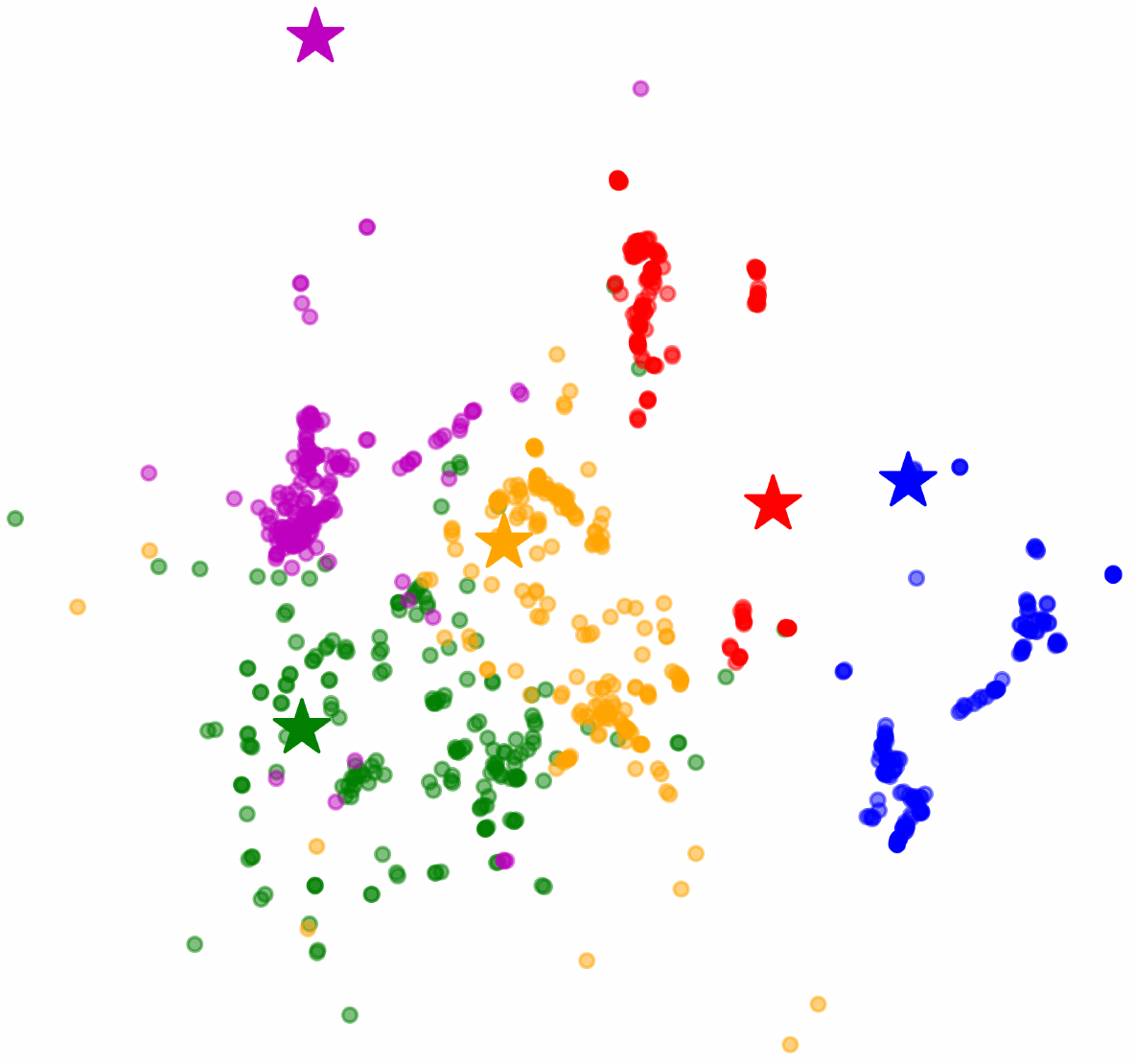}
		\label{all_qu}}
	\subfigure[Support and generated examples.]{
		\includegraphics[height=3.6cm,width=5.6cm]{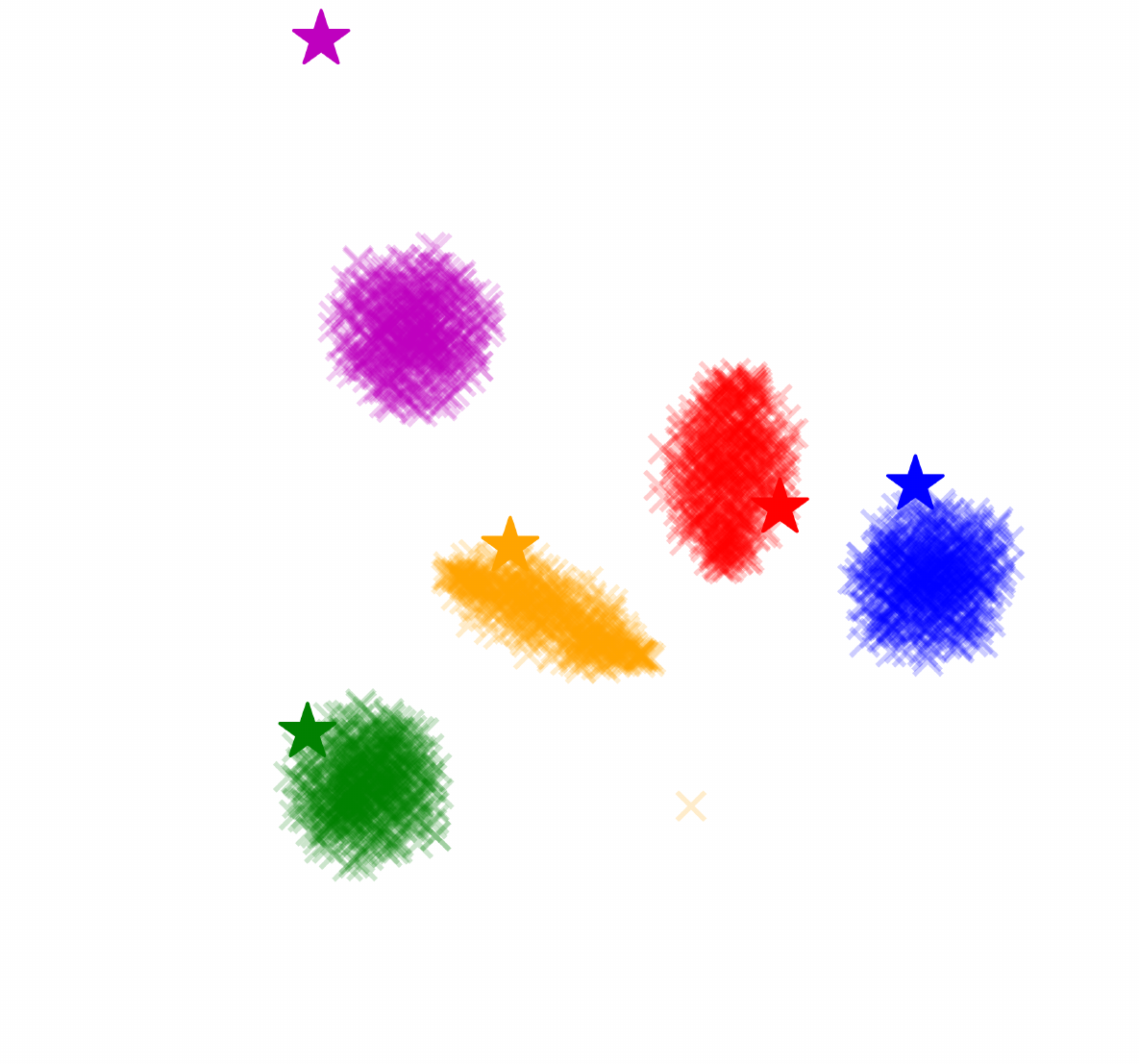}
		\label{all_sa}
	}
	\subfigure[Support, generated and query examples.]{
		\includegraphics[height=3.6cm,width=5.6cm]{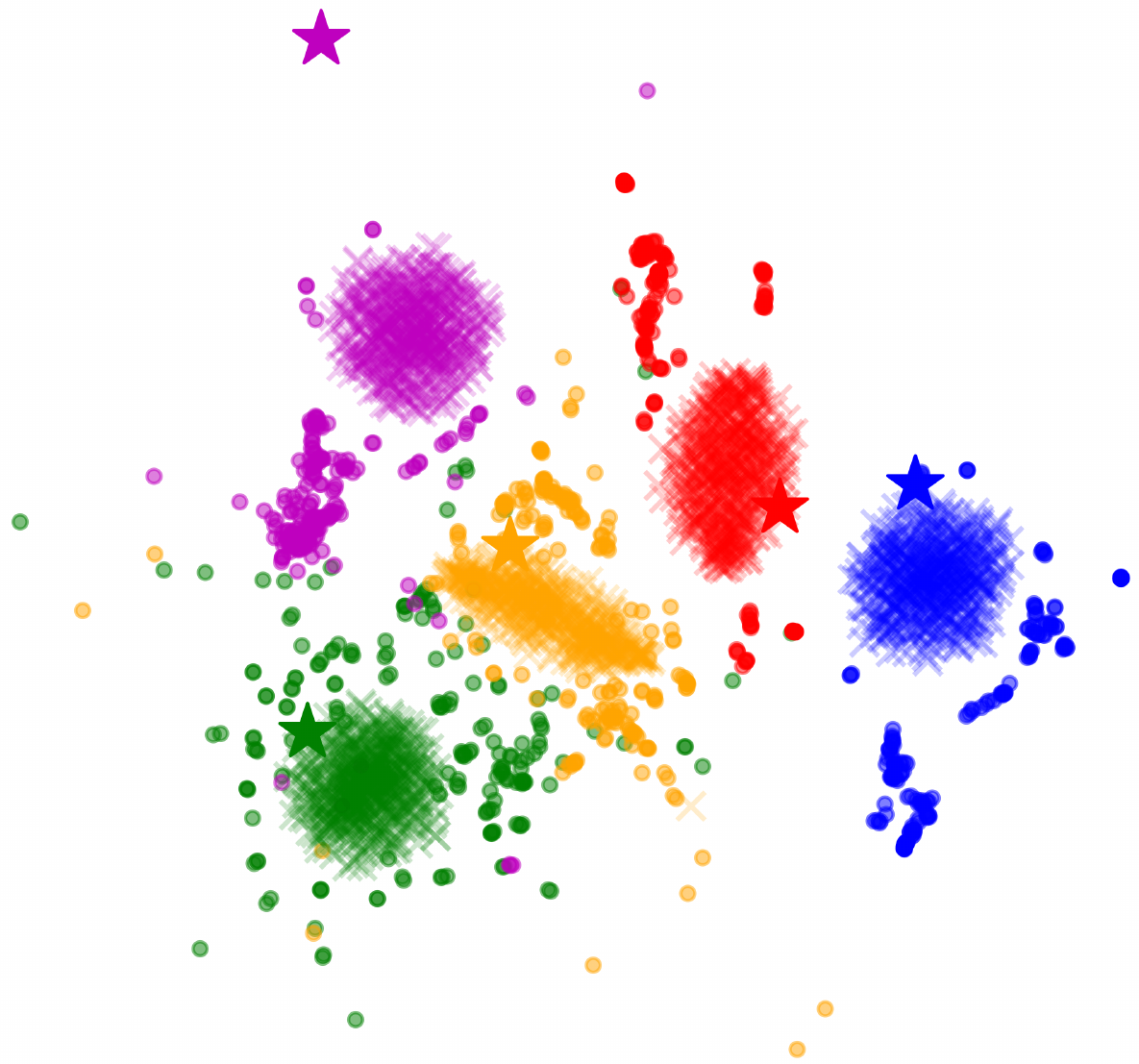}
		\label{all}
	}
	\caption{Visualization of distributions obtained by our proposed methods Way-DE/Shot-DE in 5-way 1-shot scenario. The star, dot and cross points mean support, query and generated examples respectively. Different colors denote different classes.}
	\label{visual}
\end{figure*}

\subsection{Result Analysis}
Tables \ref{text} and \ref{intent} report the experimental results for the news or review classification task and the intent detection task. Most baseline results are taken from \cite{ContrastNet} and the top 2 results are highlighted in bold. 

\subsubsection{News or Review Classification}
From Table \ref{text}, we can make the following observations. (1) Our proposed Way-DE and Shot-DE methods perform much better than other baselines in most cases, and achieve the best performance in average. Specifically, in the 1-shot and 5-shot scenarios, from the average perspective, our proposed methods improves upon existing methods by 1.2\%-32.0\% and 2.1\%-36.3\%. The reason is that Way-DE and Shot-DE strategies can accurately estimate the distribution and generate available samples, thus providing strong supervision for training the classification model.  (2) Some powerful baselines like ContrastNet and TPN also perform well in most cases. The reason is that they use a large amount of unlabeled data in target domain. While just leveraging very limited queries for each episode, our approaches still outperform them significantly, which further demonstrates the superiority of our proposed methods.

\subsubsection{Intent Detection}	
From Table \ref{intent}, it is easy to find that: (1) Compared with these latest methods, our proposed methods can achieve very competitive performance. Specifically, on the Liu57 dataset, the average accuracy of Way-DE and Shot-DE methods is greater than 90\% and 95\% in 1-shot and 5-shot scenarios, which outperforms other algorithms greatly. These improvements indicate that estimating distribution using queries and then sampling from distribution can effectively mitigate the data scarcity issue in few-shot learning. (2) Limited by the number of queries, the improvement of our proposed methods is affected, but they still perform better than other baselines, which validates the effectiveness of the proposed strategies.

\subsection{Comparison of Distribution Estimation Strategies}
In order to deeply explore the disparity of different distribution estimation methods, we conduct a series of experiments under various conditions. DC \cite{freelunch} is the distribution calibration method, which transfers the statistics from seen classes to unseen classes. DC-DE is our modified method, which considers the statistics of seen classes and query data simultaneously. Way-DE and Shot-DE are our proposed distribution estimation methods by just utilizing query samples. The results are shown in Tables \ref{text} and \ref{intent}. We can observe that Way-DE and Shot-DE perform much better than DC and DC-DE, and their results are very similar. The reason is that our proposed Way-DE and Shot-DE employ unlabeled query samples instead of adequate samples from seen
classes, thus circumventing the possible adverse impact caused by transferring the statistics of seen classes. As Way-DE and Shot-DE have the same mean, their results tend to be consistent. In addition, DC-DE outperforms DC, but not as well as Way-DE and Shot-DE, which indicates that combining the distribution information of seen classes and query data may not bring further improvement, even may be detrimental in most cases.

\subsection{Visualization}
To show what the estimated distribution looks like, we use t-SNE \cite{van2008visualizing} to visualize the distributions. To be convenient to observe the real distributions, we use 200 unlabeled query examples and 500 generated examples per class from the Liu57 dataset under 5-way 1-shot setting. Note that in the 1-shot scenario, Way-DE and Shot-DE are equivalent in principle. Figure \ref{all_qu} shows the original support and query examples. Figure \ref{all_sa} shows the support and generated examples. Figure \ref{all} shows the support, generated and query examples, which provides a comprehensive distribution representation. We have the following observations: (1) In Figure \ref{all_qu}, due to the scarcity of support set, only one example in this case, the support set is more likely mismatch with the query set. (2) In Figure \ref{all_sa}, by leveraging several query examples, the generated examples can better calibrate the real distribution, thus avoiding some support examples locate in the margin of the distribution. (3) In Figure \ref{all}, the generated examples overlap largely with the query features, which indicates our distribution estimation is accurate and reasonable. Therefore, training and testing with these examples can boost the performance effectively.

\section{Conclusion}
In this paper, we propose two simple and sweet distribution estimation methods to deal with the few-shot text classification task. By utilizing top nearest queries to calibrate the data distribution and generate more informative samples according to the estimated distribution, the proposed methods can avoid the potential negative impact caused by transferring from irrelevant seen classes, thus obtaining a more powerful classifier model for few-shot text classification. Extensive experimental results on four news or review classification datasets and four intent detection datasets show that our proposed Way-DE and Shot-DE outperform the state-of-the-art methods by a large margin. In future work, we plan to further investigate the theoretical underpinnings of our proposed strategies, and extend the strategies to deal with the multi-label few-shot text classification task.

\section{Acknowledgments}
The authors are grateful to the anonymous reviewers for their valuable comments. This work was supported by National Natural Science Foundation of China (No. 62106035, 62206038, 61972065) and Fundamental Research Funds for the Central Universities (No. DUT20RC(3)040, DUT20RC(3)066), and supported in part by Key Research Project of Zhejiang Lab (No. 2022PI0AC01) and National Key Research and Development Program of China (2022YFB4500300). We also would like to thank Dalian Ascend AI Computing Center and Dalian Ascend AI Ecosystem Innovation Center for providing inclusive computing power and technical support.

\bibliography{aaai23}

\end{document}